\documentclass[conference]{IEEEtran}
\IEEEoverridecommandlockouts

\usepackage{cite}
\usepackage{amsmath,amssymb,amsfonts}
\usepackage{graphicx}
\usepackage{textcomp}
\usepackage{hyperref}
\usepackage{xcolor}
\usepackage[caption=false,font=footnotesize]{subfig}

\def\BibTeX{{\rm B\kern-.05em{\sc i\kern-.025em b}\kern-.08em
    T\kern-.1667em\lower.7ex\hbox{E}\kern-.125emX}}

\usepackage[acronym,shortcuts]{glossaries}
\newacronym{CNN}{CNN}{convolutional neural network}
\newacronym{SPV}{SPV}{simulated prosthetic vision}
\newacronym{DCT}{DCT}{discrete cosine transform}
\glsdisablehyper

\begin{document}

\title{SymbolSight: Minimizing Inter-Symbol Interference for Reading with Prosthetic Vision}

\author{Jasmine~Lesner$^{1,*}$, Michael~Beyeler$^{1,2}$%
\thanks{$^{*}$Corresponding author: \url{jlesner@ucsb.edu}}%
\thanks{$^{1}$J.L.~and M.B.~are with the Department of Computer Science, University of California, Santa Barbara, CA 93106, USA.}
\thanks{$^{2}$M.B.~is also with the Department of Psychological \& Brain Sciences, University of California, Santa Barbara, CA 93106, USA.}
}

\maketitle

\begin{abstract}
Retinal prostheses restore limited visual perception, but low spatial resolution and temporal persistence make reading difficult. In sequential letter presentation, the afterimage of one symbol can interfere with perception of the next, leading to systematic recognition errors. Rather than relying on future hardware improvements, we investigate whether optimizing the visual symbols themselves can mitigate this temporal interference. We present \textsc{SymbolSight}, a computational framework that selects symbol-to-letter mappings to minimize confusion among frequently adjacent letters. Using simulated prosthetic vision (SPV) and a neural proxy observer, we estimate pairwise symbol confusability and optimize assignments using language-specific bigram statistics. Across simulations in Arabic, Bulgarian, and English, the resulting heterogeneous symbol sets reduced predicted confusion by a median factor of 22 relative to native alphabets. 
These results suggest that standard typography is poorly matched to serial, low-bandwidth prosthetic vision and demonstrate how computational modeling can narrow the design space of visual encodings, identifying high-potential candidates for future psychophysical and clinical evaluation rather than predicting present-day clinical reading performance directly.
\end{abstract}



\section{Introduction}

Retinal prostheses provide a crucial restorative option for individuals with degenerative retinal diseases, but functional tasks like reading remain difficult because the effective spatiotemporal channel is severely constrained.
To date, PRIMA is the only retinal implant with published evidence of form vision~\cite{holz_subretinal_2025} in the sense of reliably perceiving structured spatial patterns. 
At the same time, earlier devices such as Alpha-IMS and Argus II showed that letter-level reading is not categorically impossible for all users~\cite{zrenner_subretinal_2011,cruz_argus_2013}.
With training and careful stimulus presentation, some participants could identify isolated characters presented one symbol at a time.

\begin{figure}[htbp]
    \centering
    \includegraphics[
        width=\linewidth,
        trim=11.3cm .3cm 10.3cm .3cm, 
        clip
    ]{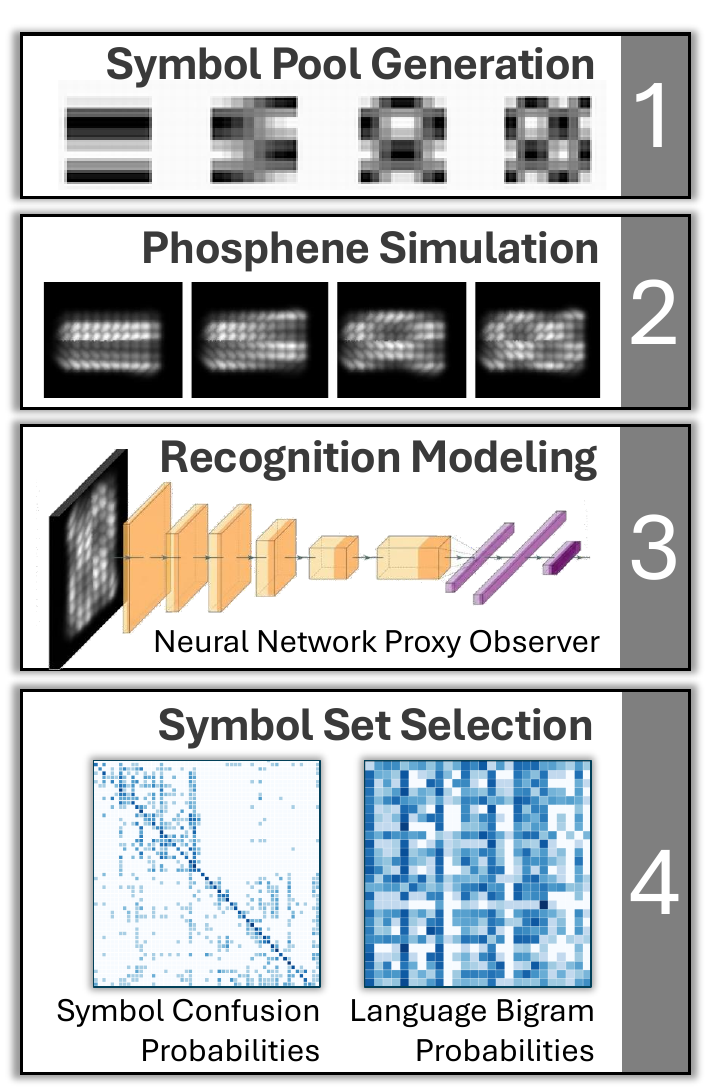}

    \caption{
        \textsc{SymbolSight} pipeline: candidate symbols undergo phosphene simulation and recognition modeling, then are assigned to letters based on confusion probabilities and bigram statistics.
    }
    \label{fig:ssight_overview}
\end{figure}

These mixed outcomes point to two interacting bottlenecks that limit sequential reading in practice. First, spatial nonlinearities can break the assumption that multi-electrode stimulation produces a percept that is the linear sum of single-electrode percepts~\cite{beyeler_learning_2017,hou_axonal_2024}. This complicates any attempt to render conventional letterforms, since increasing spatial detail by recruiting more electrodes does not necessarily yield predictable, compositional structure at the perceptual level~\cite{christie_sequential_2022}. 
Second, temporal nonlinearities introduce inter-symbol interference. 
Percepts can persist and fade over hundreds of milliseconds or longer~\cite{horsager_temporal_2011,perez_fornos_temporal_2012}, so the afterimage of one symbol can distort the next. In this regime, the discriminability of a symbol depends not only on its own geometry but also on the preceding symbol.

Prior work on reading in bionic vision has largely focused on modeling and accommodating known distortions~\cite{fornos_reading_2011,vurro_simulation_2014,paraskevoudi_full_2021}. 
Complementary work in applied perception and display design has explored how symbols and typefaces can remain legible under degraded viewing conditions~\cite{wickens2021engineering, garvey1998development}. 
These efforts provide important foundations, but they typically assume a fixed alphabet and attempt to recover it through better hardware, better stimulation, or better decoding.
In contrast, here we ask a different question: \emph{given that the channel is low-bandwidth and exhibits spatial and temporal nonlinearities, what symbol set should we transmit in the first place?}

To test this, we present \textsc{SymbolSight} (Fig.~\ref{fig:ssight_overview}), a framework that optimizes the symbol-to-letter mapping for specific languages. The pipeline consists of: (1) generating a diverse pool of candidate glyphs; (2) simulating prosthetic spatial and temporal distortions; (3) estimating pairwise confusability using a neural network as a consistent proxy observer; and (4) assigning symbols to letters to minimize the expected confusion cost based on language-specific letter transition probabilities.
Our primary contributions are:
\begin{itemize}
    \item We introduce a modular computational pipeline that combines \ac{SPV}, a neural proxy observer, and combinatorial optimization. This approach allows us to navigate the vast search space of potential symbol mappings, filtering for candidates that maximize distinguishability prior to costly human-subject testing.
    \item We formulate a language-aware cost function that specifically targets temporal persistence, a key bottleneck in sequential reading. By weighting symbol confusability against language-specific bigram frequencies, we minimize perceptual interference for the most frequent letter transitions rather than treating all errors equally.
    \item We demonstrate that optimal legibility in prosthetic vision may require abandoning typographic cohesion. In simulations across Arabic, Bulgarian, and English, we find that mixed pools containing diverse glyphs (e.g., Latin, Braille, \ac{DCT} basis functions) reduce predicted confusion by a median factor of 22 compared to native alphabets. This finding challenges the assumption that standard letters are the appropriate baseline for prosthetic interfaces.
\end{itemize}

\section{Methods}

We render high-contrast symbols and present them one at a time, consistent with standard \ac{SPV} paradigms. 
Due to implant limitations, symbols are shown sequentially rather than in parallel. 
Depending on electrode placement and axonal activation, users may perceive small focal spots, elongated streaks, or irregular shapes rather than uniform patterns~\cite{beyeler2019model,sinclair_appearance_2016}. 
Our goal is to model how these spatiotemporal distortions affect symbol discrimination and to optimize symbol assignments accordingly, not to reproduce the full gaze-contingent reading strategies used by current implant recipients.

\subsection{Simulated Prosthetic Vision}
We use pulse2percept~\cite{beyeler2017pulse2percept,granley_computational_2021} to simulate a $16 \times 16$ electrode grid, a resolution commonly used in prior \ac{SPV} studies. While higher-density arrays may provide additional spatial detail, current spread and axonal activation often limit effective resolution~\cite{beyeler_learning_2017}.

\begin{figure*}[htbp]
    \centering
    \subfloat{\includegraphics[width=0.329\textwidth]{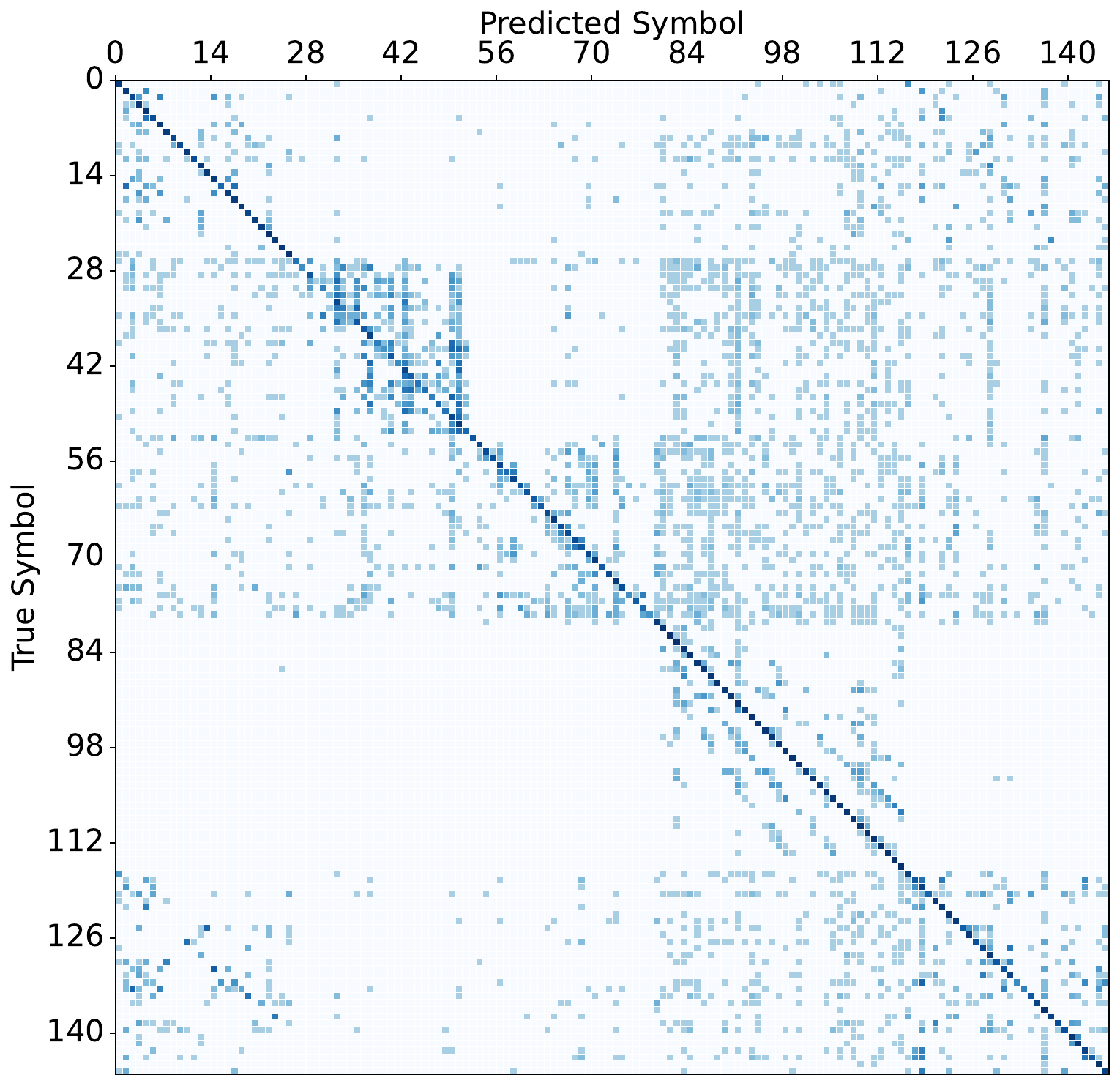}}
    \subfloat{\includegraphics[width=0.335\textwidth]{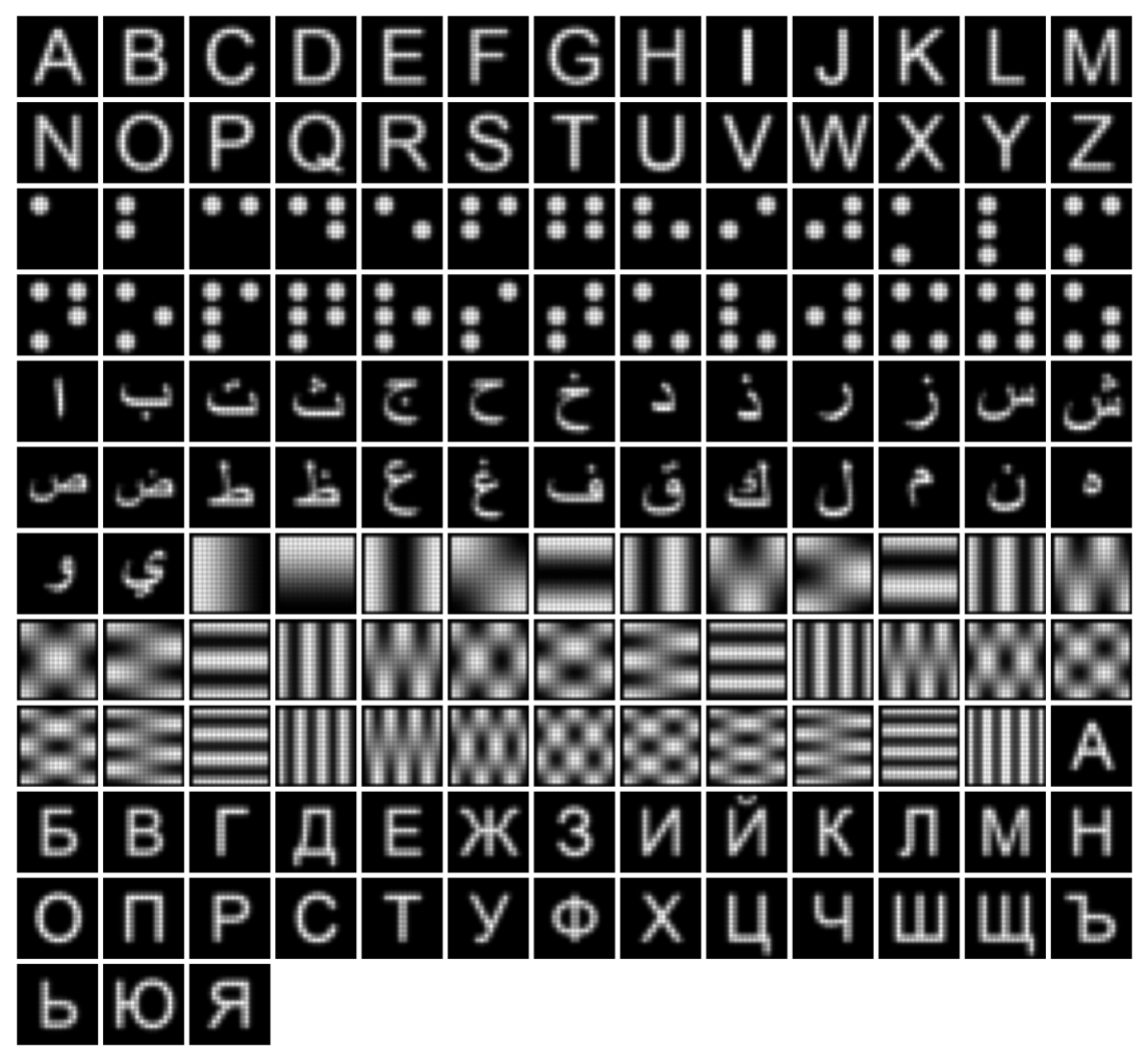}}
    \subfloat{\includegraphics[width=0.335\textwidth]{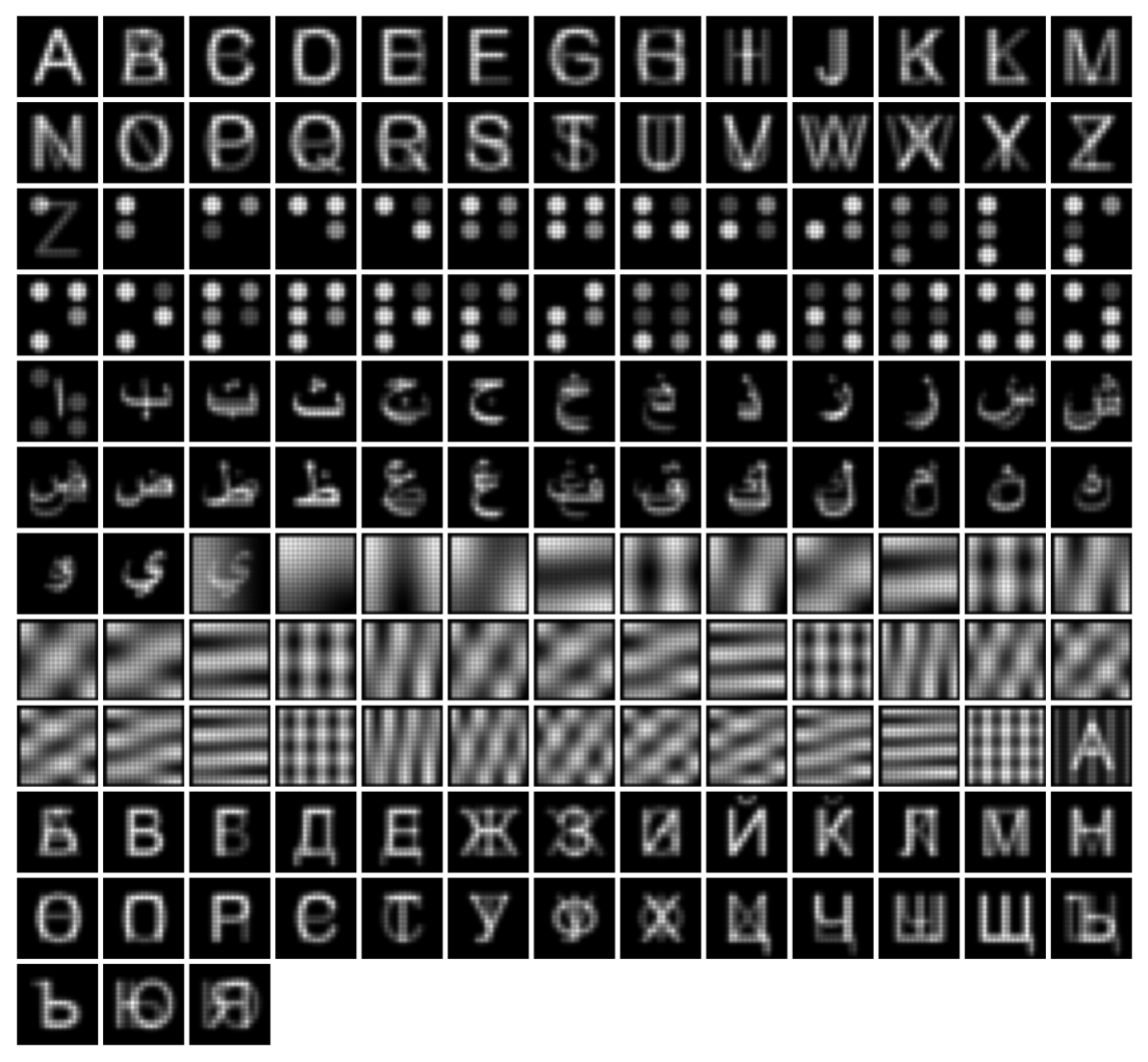}}\\[1ex]
    \subfloat{\includegraphics[width=0.329\textwidth]{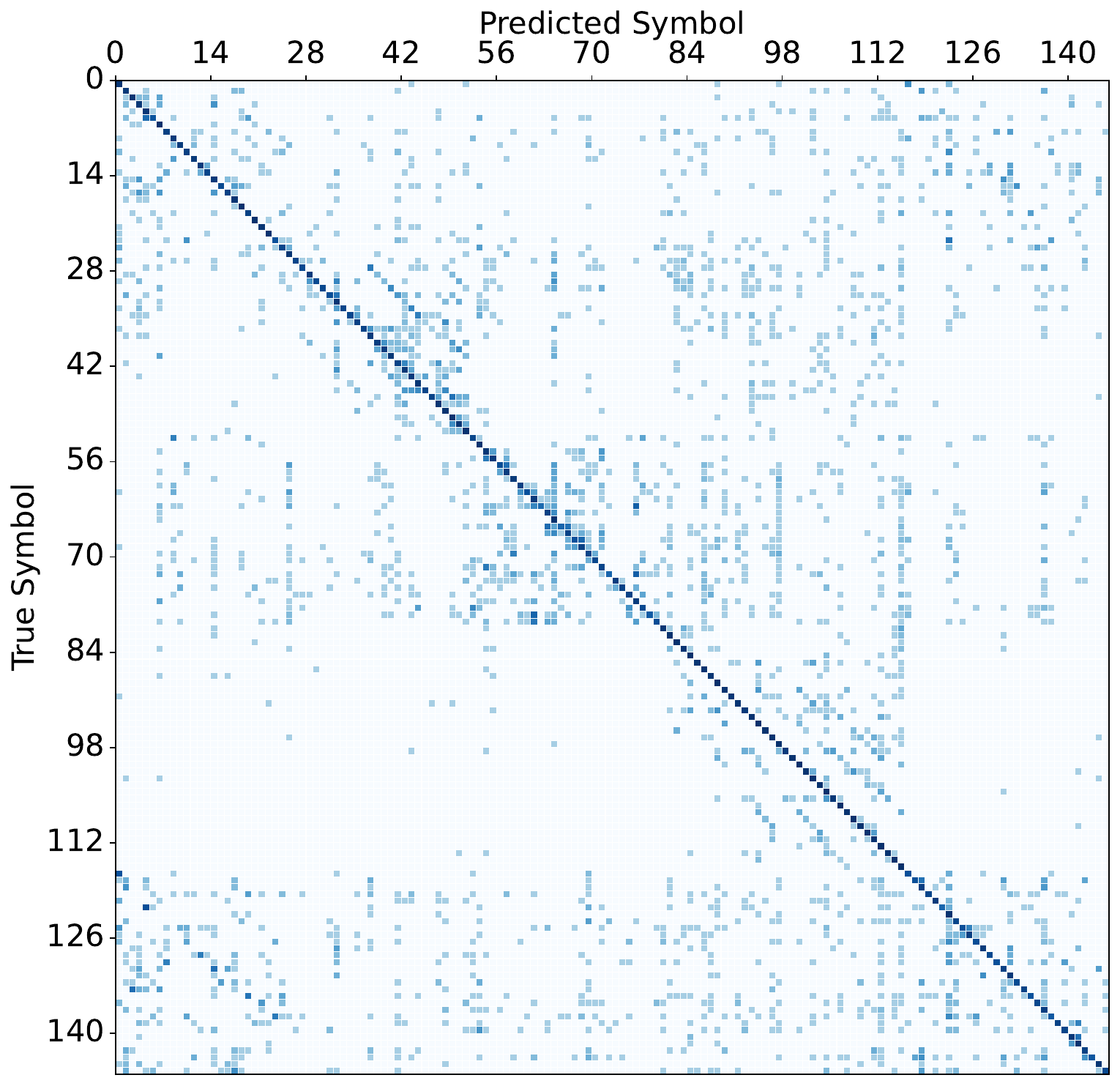}}
    \subfloat{\includegraphics[width=0.335\textwidth]{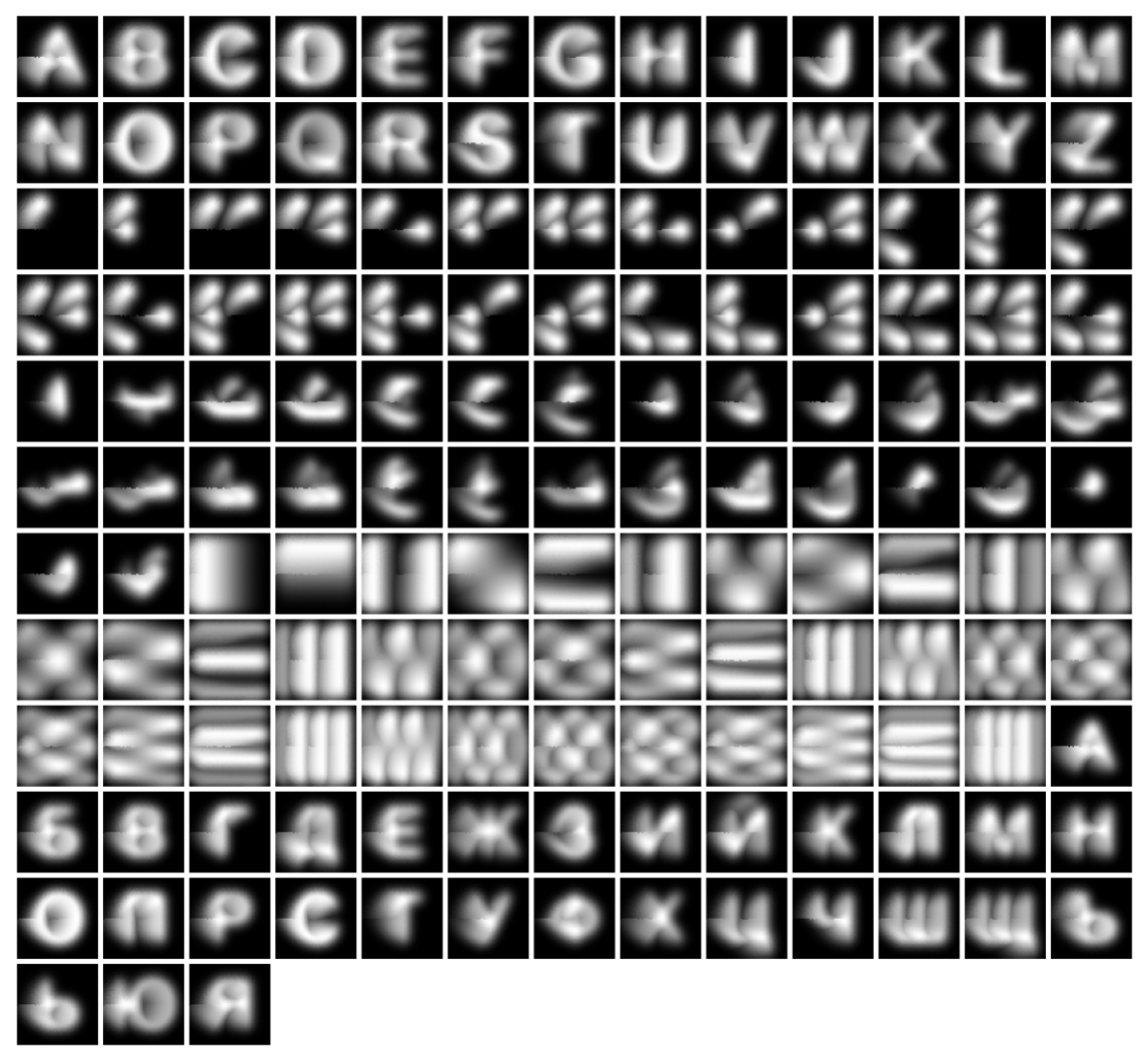}}
    \subfloat{\includegraphics[width=0.335\textwidth]{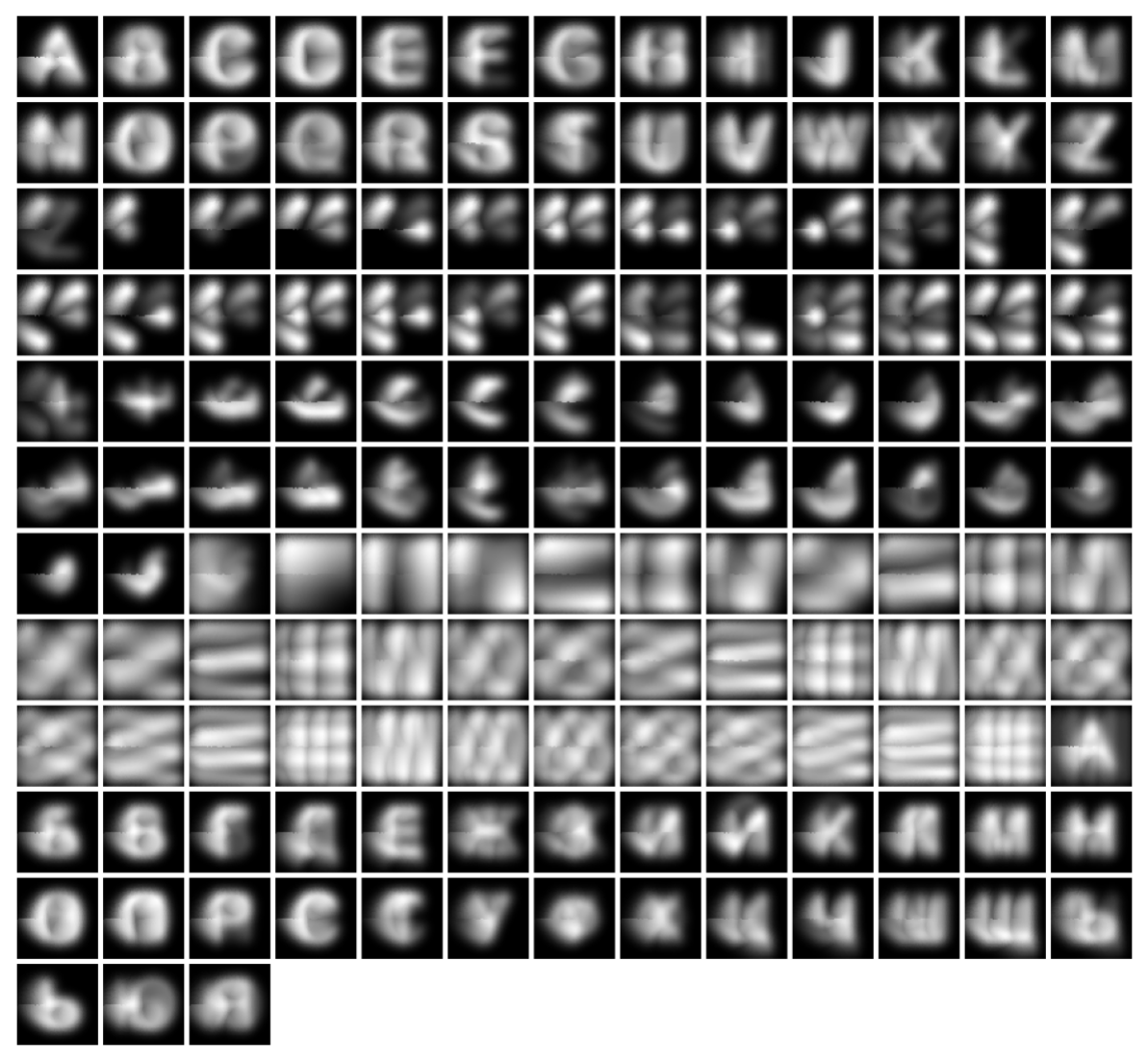}}\\[1ex]
    \subfloat{\includegraphics[width=0.329\textwidth]{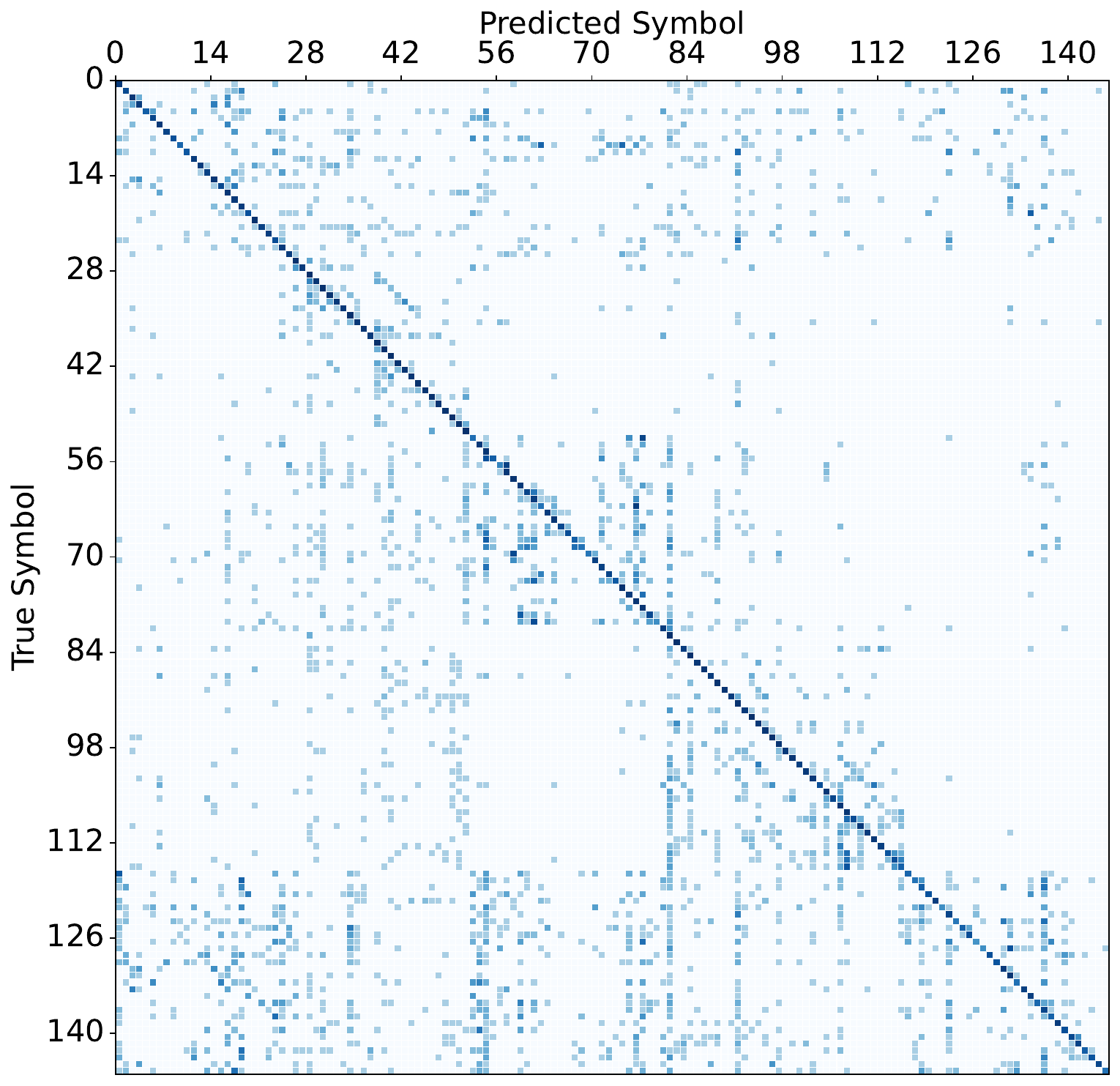}}
    \subfloat{\includegraphics[width=0.335\textwidth]{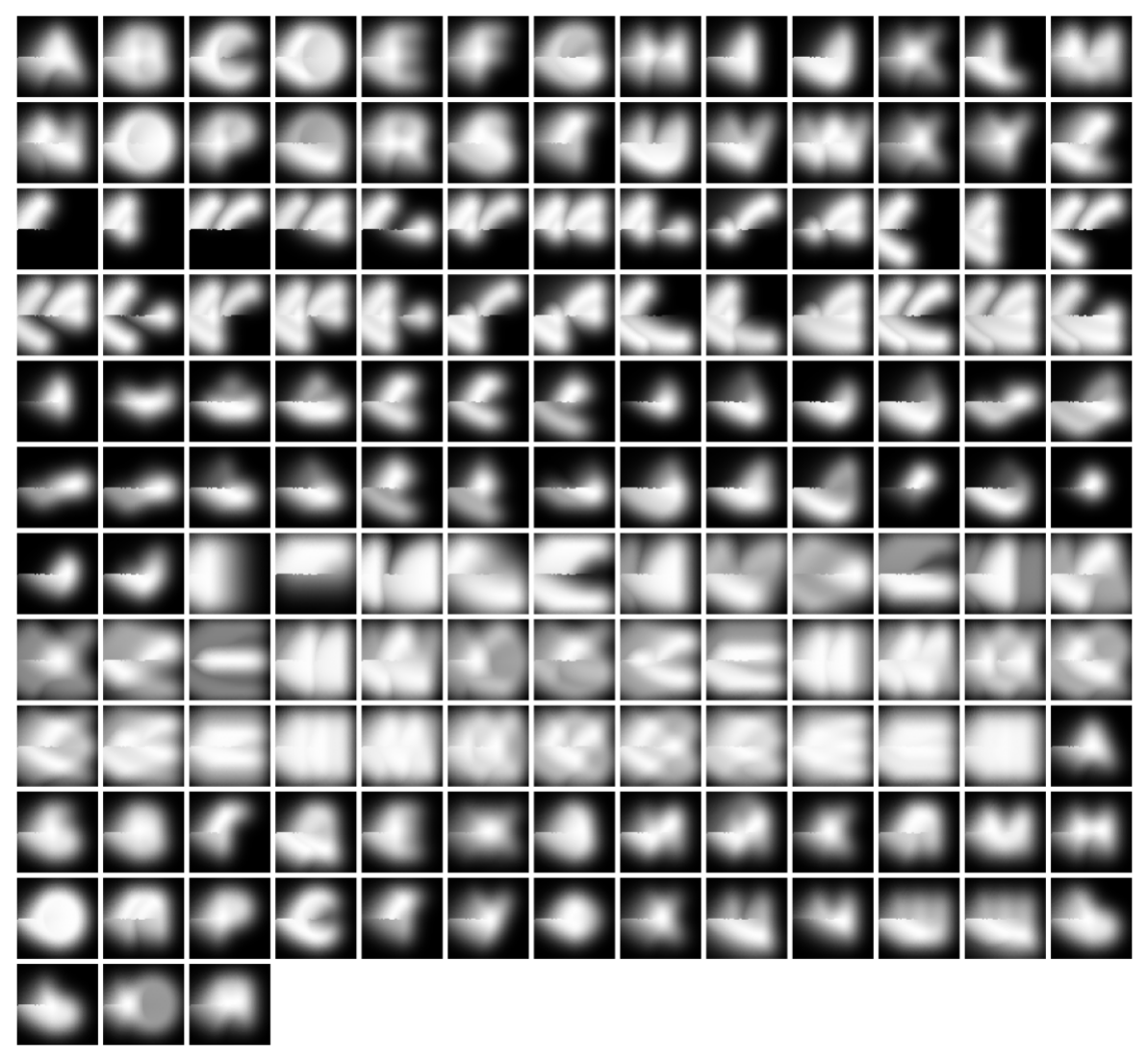}}
    \subfloat{\includegraphics[width=0.335\textwidth]{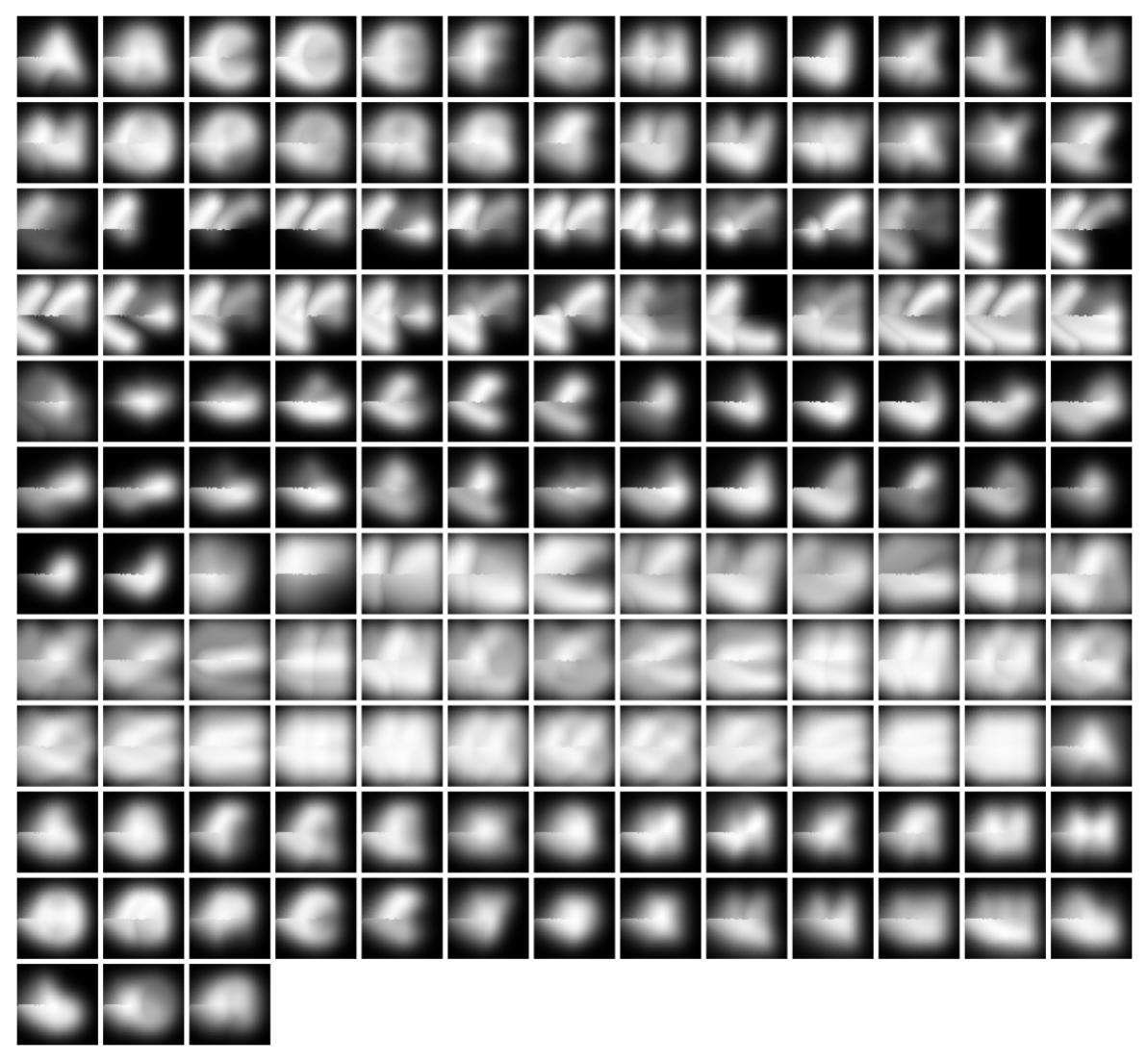}}
    \caption{
        \textbf{Top row}: low distortion; 
        \textbf{middle row}: medium distortion; 
        \textbf{bottom row}: high distortion.
        \textbf{Middle column}: 146 symbols with spatial distortion (Latin 0--25, Braille 26--51, Arabic 52--79, \ac{DCT} 80--115, Cyrillic 116--145). 
        \textbf{Right column}: example of temporal distortion showing perceptual residue when symbols are presented sequentially.
        \textbf{Left column}: symbol confusion probability heatmaps from fine-tuned neural networks evaluated on a held out validation dataset. Darker cells indicate higher confusion; white indicates near-zero.
        Our proxy observer networks were trained on a symbol pool of 146 glyphs comprising 26 Latin letters, 26 Braille characters, 28 Arabic letters, 36 \ac{DCT} basis patterns, and 30 Cyrillic letters. Each symbol was rendered at $64 \times 64$ pixels and processed through four distortion conditions using the pulse2percept library~\cite{beyeler2017pulse2percept}: undistorted, and three prosthetic vision simulations on a $16 \times 16$ electrode grid with phosphene sizes $\rho \in \{100, 300, 500\}\,\mu\mathrm{m}$ and axon streak parameters $\lambda \in \{0, 1000, 5000\}\,\mu\mathrm{m}$. For each distortion level, a separate MobileNetV3Large~\cite{howard2019searching} classifier was trained with ImageNet-pretrained weights frozen and a custom head consisting of global average pooling, 20\% dropout, and a 146-way softmax output layer with L1-L2 regularization ($\ell_1 = 0.001$, $\ell_2 = 0.02$). Training used the Adam optimizer with learning rate of $10^{-4}$, batch size of 64, and MixUp augmentation~\cite{zhang2018mixup} with $\alpha = \beta = 2.0$ to generate 500 augmented training samples per symbol (73,000 images) and 100 validation samples per symbol (14,600 images). Early stopping monitored validation loss with patience of 20 epochs, and learning rate was reduced by half after 3 epochs without improvement. To match MobileNetV3Large's input requirements, images were resized to $224 \times 224$ RGB. The train and validation sets contained all 146 symbol classes with independently augmented samples, ensuring no leakage between splits.
    }
    \label{fig:training_confusion_symbol_sets}
\end{figure*}

\textbf{Spatial Distortion}.
We simulate three levels of spatial distortion (shown top to bottom in the middle column of Fig.~\ref{fig:training_confusion_symbol_sets}) following Han et al.~\cite{han2021deep}:
\emph{Low distortion}: no axonal stimulation ($\rho = 100\,\mu\mathrm{m}$, $\lambda = 0\,\mu\mathrm{m}$);
\emph{Medium distortion}: intermediate axonal stimulation ($\rho = 300\,\mu\mathrm{m}$, $\lambda = 1000\,\mu\mathrm{m}$);
\emph{High distortion}: strong axonal stimulation ($\rho = 500\,\mu\mathrm{m}$, $\lambda = 5000\,\mu\mathrm{m}$).

\textbf{Temporal Distortion}.
In sequential prosthetic reading, the percept of a current symbol $x_t$ is degraded by the visual persistence of the preceding symbol $x_{t-1}$. While biological persistence involves complex, nonlinear decay dynamics~\cite{hou_predicting_2024,kasowski_simulated_2025}, the dominant recognition bottleneck is inter-symbol interference, where the fading afterimage of the previous symbol obscures the next.

To capture this effect without full time-domain simulation, we approximate temporal interference as linear superposition. We use MixUp augmentation~\cite{zhang2018mixup} to generate convex combinations of symbol pairs:
\begin{equation}
\hat{x} = \gamma x_i + (1 - \gamma) x_j, \quad \gamma \sim \mathrm{Beta}(2.0, 2.0),
\end{equation}
where $x_i$ and $x_j$ denote pixel intensity vectors for the current and preceding symbols. Unlike standard MixUp, we set $\alpha = \beta = 2.0$ to emphasize high-overlap cases ($\gamma \approx 0.5$), reflecting strong temporal ghosting. 
This abstraction captures the code-design problem posed by persistence, but does not model active scanning, phosphene remapping, or patient-specific temporal adaptation.

\begin{figure*}[htbp]
    \centering
    \subfloat{\includegraphics[width=0.333\textwidth]{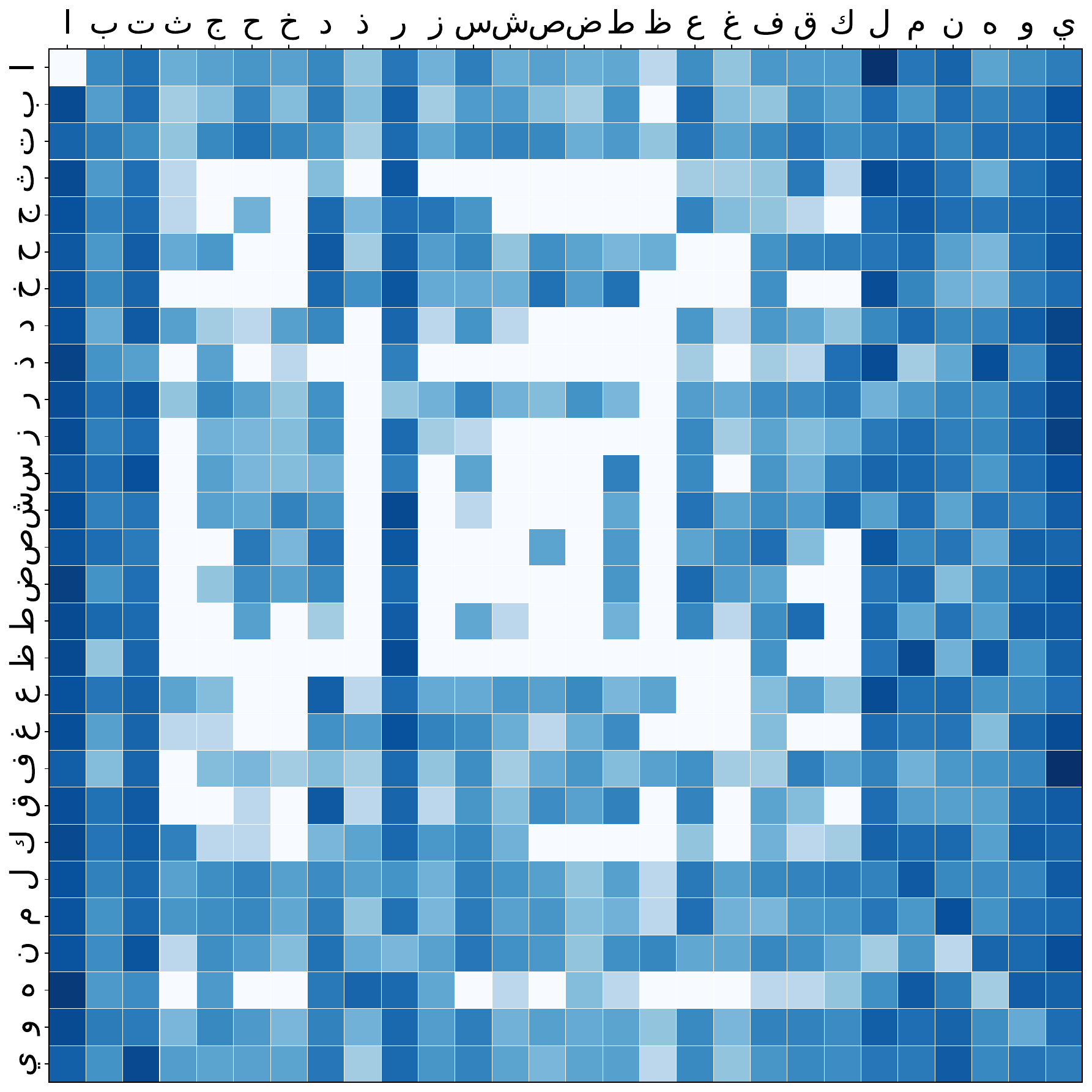}}%
    \hfill
    \subfloat{\includegraphics[width=0.333\textwidth]{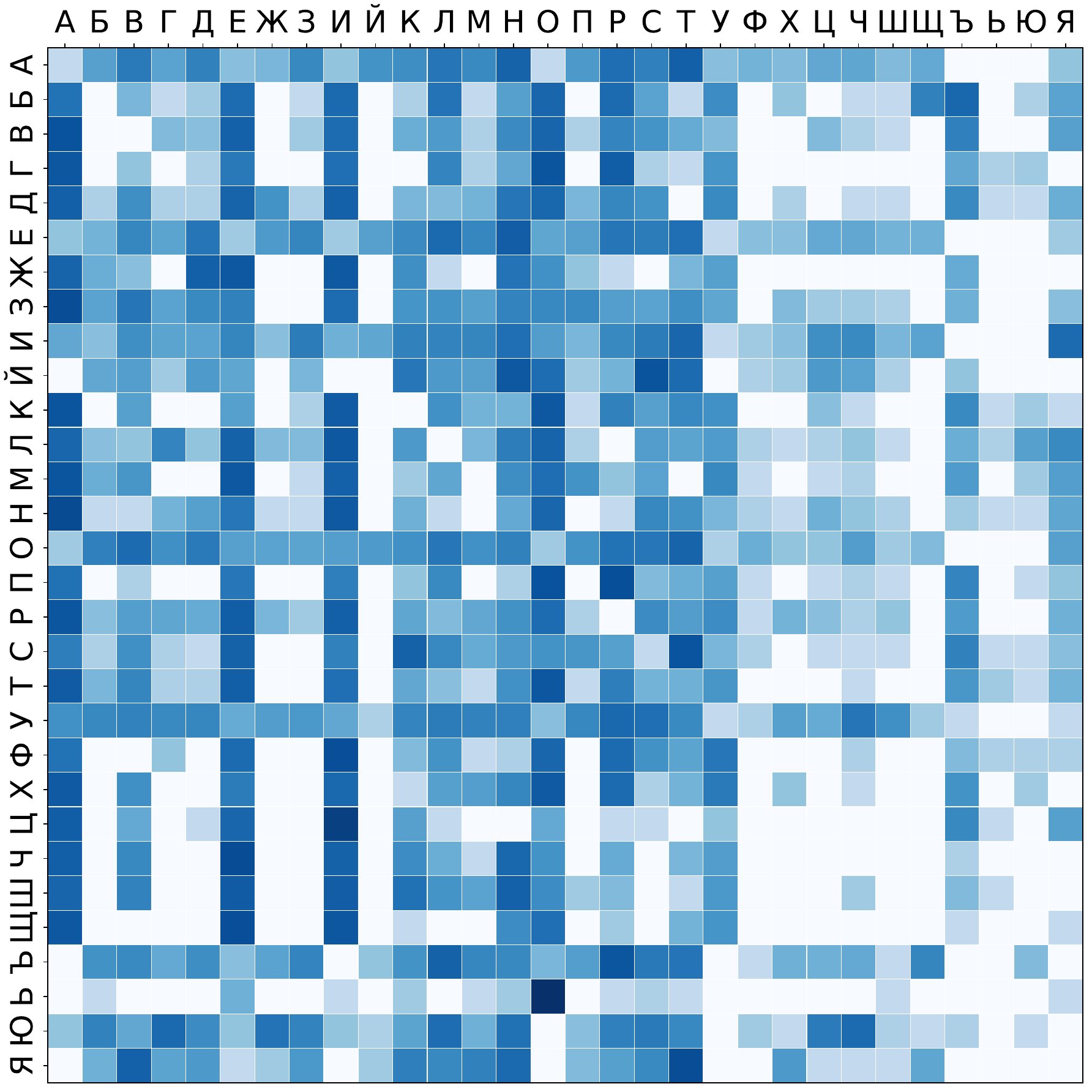}}%
    \hfill
    \subfloat{\includegraphics[width=0.333\textwidth]{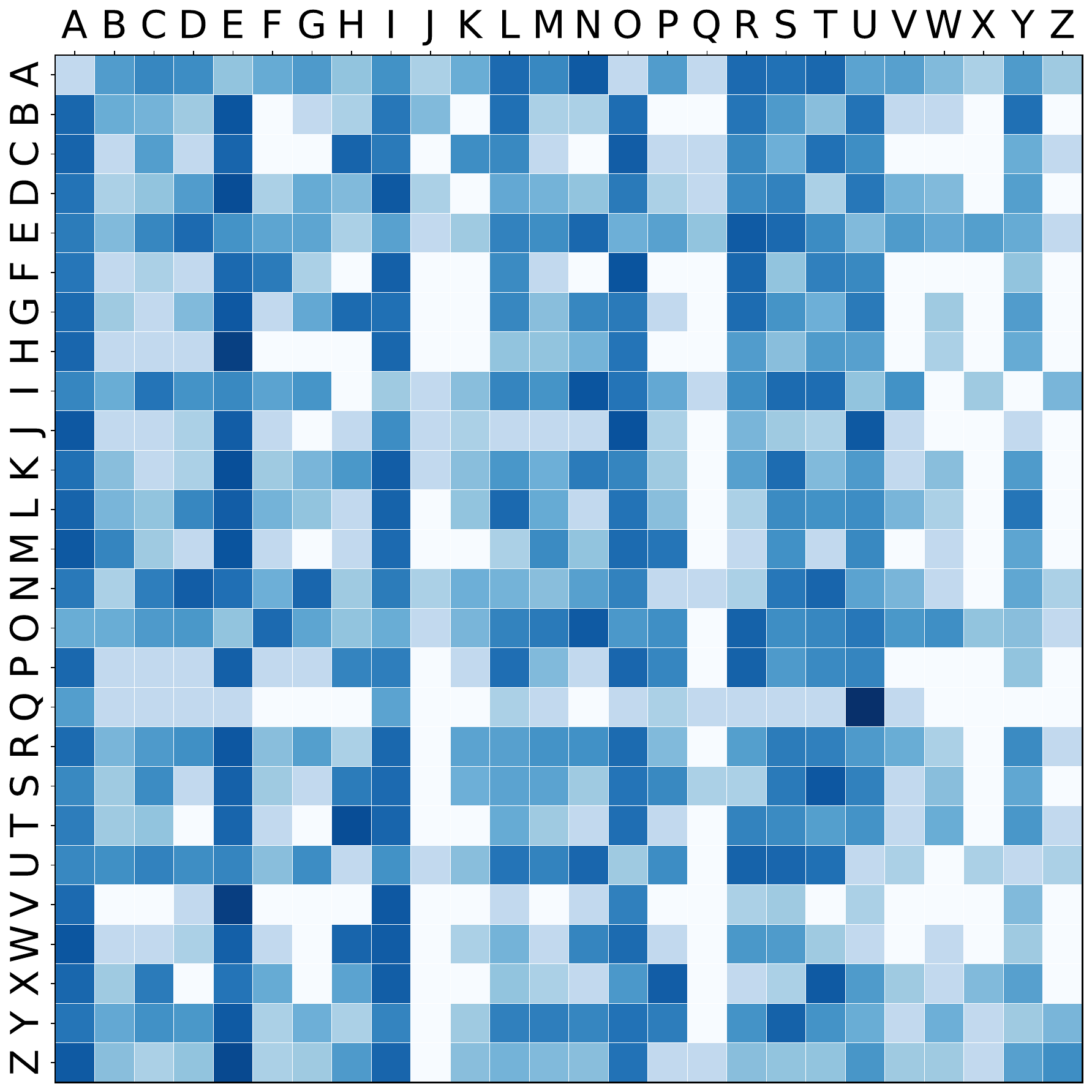}}
    \caption{\textbf{Letter transition probabilities across three languages.} 
        Heatmaps show $P(L_{n+1} \mid L_n)$ for \textbf{Arabic} (left), \textbf{Bulgarian} (middle), and \textbf{English} (right). 
        The vertical axis is the leading (current) letter; the horizontal axis is the following (next) letter. 
        Darker cells indicate higher probability. 
        Language-specific bigram probabilities were estimated from November 2023 Wikipedia database dumps~\cite{wikimedia2023} for Arabic, Bulgarian, and English. For each language, we processed the first 100,000 Wikipedia documents. English text was case-folded and filtered to the 26 Latin letters using regex word extraction. Bulgarian text was Unicode NFC-normalized and case-folded across its 30-letter Cyrillic alphabet, with 
        the accented variant (Cyrillic~I with grave accent) mapped to plain Cyrillic~I. Arabic text underwent NFKD decomposition with diacritical marks stripped, \emph{tatweel} removed, \emph{alif maqsura} normalized to \emph{yeh}, \emph{ta marbuta} mapped to \emph{ta}, and standalone \emph{hamza} treated as a word boundary, yielding a 28-letter alphabet. Documents with fewer than 500 valid letters were excluded to ensure reliable statistics. The resulting co-occurrence matrices were normalized to estimate conditional bigram probabilities $P(L_{n+1}\mid L_n)$ used in the assignment optimization. For visualization, each row of the heatmap is normalized to sum to~$1$.
        }
    \label{fig:letter_transition_probabilities}
\end{figure*}

\subsection{Simulated Observer}

We quantify how overlapping percepts affect recognition using a \ac{CNN} pre-trained on ImageNet as a simulated observer. We chose MobileNetV3Large~\cite{howard2019searching} for computational efficiency. Under each distortion condition, we fine-tune a separate classifier and estimate pairwise confusion probabilities across the full symbol pool.

While \acp{CNN} do not replicate human psychophysics in detail and may differ in decision strategies for specific stimuli, prior work has shown that task-optimized deep networks operating on simulated prosthetic vision exhibit strong correspondence with human performance across multiple functional visual tasks, including letter and symbol recognition under phosphene distortions~\cite{skaza_deep_2025}. We do not use the network to predict absolute human error rates. Instead, the resulting confusion matrix $F$ is used to rank the relative separability of symbol pairs under overlap. This acts as a filter for structural ambiguity: if a high-capacity feature extractor cannot reliably distinguish two superimposed symbols under simulated distortion, the signal itself likely lacks sufficient information for robust discrimination by a human observer. We therefore treat the \ac{CNN}-derived confusion matrix as a consistent metric of \emph{structural distinguishability}.

\subsection{Symbol Families}

For testing, we construct a diverse pool of visual symbols drawn from established alphabets and simple synthetic patterns:
\emph{Latin Symbols}: The 26-letter alphabet used by English and many other languages.
\emph{Braille Symbols}: A tactile system of raised dots in $2 \times 3$ cells, optimized for fingertip discrimination rather than visual recognition.
\emph{Arabic Symbols}: The 28-letter Arabic script with cursive, connected letterforms.
\emph{\ac{DCT} Symbols}: Basis functions from the two-dimensional discrete cosine transform~\cite{ahmed1974discrete}, sinusoidal gratings that efficiently capture fundamental visual features.
\emph{Cyrillic Symbols}: The 30-letter Bulgarian alphabet, representative of the broader Cyrillic family. Adapted from Greek, it includes letters shared with Latin (e.g., A, O, T) alongside distinctly Slavic forms (e.g., Zhe, Sha, Yu, Ya).

\subsection{Symbol Assignment Optimization}

We combine two empirical inputs: (i) a symbol confusion matrix $F \in \mathbb{R}^{S \times S}$ measured under simulated prosthetic distortions, and (ii) a language-specific letter bigram matrix $C \in \mathbb{R}^{L \times L}$. We seek a mapping $\pi$ from letters to symbols that minimizes expected confusion between successive letters:
\begin{equation}
\mathrm{Cost}
= \sum_{i=0}^{L-1} \sum_{\substack{j=0 \\ j \neq i}}^{L-1}
C_{i,j} \, F_{\pi(i), \pi(j)}
\label{eq:confusion_cost}
\end{equation}

This cost increases when frequently co-occurring letters are assigned to visually confusable symbols. Minimizing Eq.~\ref{eq:confusion_cost} reduces expected reading errors due to temporal interference. The formulation treats all letter confusions equally and is intended as a baseline; future extensions could incorporate semantic or task-dependent error weights.

We solve the assignment using the Hungarian algorithm~\cite{kuhn1955hungarian} for the reported experiments. To isolate the effect of optimization from symbol diversity alone, we compare \textsc{SymbolSight} against two baselines: \emph{Native} alphabets and \emph{Randomized} assignments drawn from the full symbol pool while ignoring $F$ and $C$. We evaluate 201 random seeds to estimate both mean performance and optimistic bounds. For each candidate alphabet, we renormalize predicted symbol distributions so probabilities sum to one over the retained symbol set before computing the cost in Eq.~\ref{eq:confusion_cost}.

\subsection{Evaluation Languages and Letter Statistics}
We evaluate the framework on Arabic, Bulgarian, and English, which differ in script topology and alphabet size. For each language, we estimate letter bigram probabilities $P(L_{n+1}\mid L_n)$ from Wikipedia text and treat these statistics as fixed, language-specific inputs defining the matrix $C$. Fig.~\ref{fig:letter_transition_probabilities} visualizes the resulting bigram matrices used in all experiments.

\section{Results}

\subsection{Symbol Confusion Structure}
Fig.~\ref{fig:training_confusion_symbol_sets} summarizes the pairwise symbol confusion probabilities estimated from our proxy observers under increasing distortion, revealing how symbol families interact under spatial and temporal interference.

First, at low distortion, the matrices exhibit a faint block structure along the diagonal, with each block corresponding to a symbol family: Latin (indices 0--25), Braille (indices 26--51), Arabic (indices 52--79), \ac{DCT} (indices 80--115), and Cyrillic (indices 116--145). This block structure indicates that symbols tend to be confused with others from the same family, reflecting within-family visual similarity. As distortion increases, this block structure progressively dissolves: family boundaries become less distinct, and off-diagonal confusion disperses more uniformly across the matrix rather than concentrating within family-specific blocks. At high distortion, the confusion pattern approaches a sparser, more scattered distribution, suggesting that prosthetic distortion obscures the fine-grained features that distinguish symbol families.

Second, cross-family confusion occurs predictably between Latin and Cyrillic symbols, reflected in off-diagonal activity linking these two blocks. This pattern persists across distortion levels, as shared letterforms such as A, E, K, M, O, and T remain visually similar even under prosthetic distortion.

Third, \ac{DCT} symbols exhibit a distinctive diagonal banding pattern within their block: because \ac{DCT} basis functions vary systematically in spatial frequency and orientation, neighboring indices often represent similar frequencies, leading to mutual confusion along diagonals. However, \ac{DCT} symbols are rarely confused with letter-based symbols from other sets, as their periodic grating structure is visually distinct from the curved and angular strokes of alphabetic characters.

Fourth, Braille symbols show elevated off-diagonal confusion at low distortion due to temporal smearing: the linear superposition of two sparse dot patterns often inadvertently creates a geometry resembling a third valid Braille symbol. Unintuitively, this systematic confusion \emph{decreases} at higher spatial distortion levels. As pulse2percept blurs individual dots into streaks, the precise combinatorial structure required for these specific misidentifications dissolves. This yields a critical insight: spatial prosthetic distortions do not uniformly degrade information but can occasionally mitigate specific temporal interference patterns.

Taken together, these observations show that symbol confusability in \ac{SPV} depends strongly on symbol family and the distortion regime, and that the resulting collision patterns are predictable and can be exploited algorithmically rather than dismissed as noise.

\begin{table}[tbp]
\centering
\caption{\textbf{Confusion Cost} (Eq.~\ref{eq:confusion_cost}). Values scaled by $10^{5}$. Lower values indicate less predicted confusion.
 }
\label{tab:confusion_cost}
\setlength{\tabcolsep}{3pt}
\begin{tabular}{lrrrr}
\hline
\textbf{Distortion} & \textbf{Native} & \textbf{Randomized} & \textbf{Randomized} & \textbf{SymbolSight} \\
 & (baseline) & (mean $\pm$ std) & (best ablation) & \textbf{Optimized} \\
\hline
\multicolumn{5}{l}{\textit{\textbf{Arabic}}} \\
Low & 715 & 607 $\pm$ 242 & 201 & \textbf{54} \\
Medium & 987 & 327 $\pm$ 194 & 124 & \textbf{45} \\
High & 1264 & 474 $\pm$ 403 & 156 & \textbf{47} \\
\hline
\multicolumn{5}{l}{\textit{\textbf{Bulgarian}}} \\
Low & 811 & 615 $\pm$ 251 & 231 & \textbf{27} \\
Medium & 422 & 308 $\pm$ 103 & 134 & \textbf{34} \\
High & 673 & 459 $\pm$ 232 & 158 & \textbf{31} \\
\hline
\multicolumn{5}{l}{\textit{\textbf{English}}} \\
Low & 700 & 646 $\pm$ 264 & 246 & \textbf{32} \\
Medium & 458 & 329 $\pm$ 176 & 133 & \textbf{24} \\
High & 563 & 475 $\pm$ 226 & 127 & \textbf{38} \\
\hline
\end{tabular}
\end{table}

\begin{table}[tbp]
\centering
\caption{\textbf{Improvement Factor} (Native / Method Cost). Higher values indicate greater improvement over native symbols.
}
\label{tab:improvement_factor}
\setlength{\tabcolsep}{3pt}
\begin{tabular}{lrrrr}
\hline
\textbf{Distortion} & \textbf{Native} & \textbf{Randomized} & \textbf{Randomized} & \textbf{SymbolSight} \\
 & (baseline) & (mean $\pm$ std) & (best ablation) & \textbf{Optimized} \\
\hline
\multicolumn{5}{l}{\textit{\textbf{Arabic}}} \\
Low & 1.0 & 1.3 $\pm$ 0.5 & 3.6 & \textbf{13.3} \\
Medium & 1.0 & 3.5 $\pm$ 1.2 & 8.0 & \textbf{21.7} \\
High & 1.0 & 3.4 $\pm$ 1.4 & 8.1 & \textbf{27.2} \\
\hline
\multicolumn{5}{l}{\textit{\textbf{Bulgarian}}} \\
Low & 1.0 & 1.5 $\pm$ 0.6 & 3.5 & \textbf{29.6} \\
Medium & 1.0 & 1.5 $\pm$ 0.5 & 3.1 & \textbf{12.5} \\
High & 1.0 & 1.8 $\pm$ 0.8 & 4.3 & \textbf{21.9} \\
\hline
\multicolumn{5}{l}{\textit{\textbf{English}}} \\
Low & 1.0 & 1.3 $\pm$ 0.5 & 2.8 & \textbf{21.6} \\
Medium & 1.0 & 1.7 $\pm$ 0.6 & 3.4 & \textbf{19.2} \\
High & 1.0 & 1.5 $\pm$ 0.7 & 4.4 & \textbf{14.7} \\
\hline
\textit{Median:} & 1.0 & --- & 3.6 & \textbf{21.6} \\
\hline
\end{tabular}
\end{table}

\subsection{Quantitative Performance Across Assignments}

To assess how these symbol collision patterns translate into reading performance, we compare native alphabets, randomized heterogeneous assignments, and optimized mappings produced by \textsc{SymbolSight}. Tables~\ref{tab:confusion_cost} and~\ref{tab:improvement_factor} show our findings.

First, \emph{native baseline costs exhibit non-monotonic behavior with respect to distortion.} Notably, native Bulgarian and English symbols yield higher predicted confusion at low distortion than at medium levels
(e.g., Bulgarian: 811 vs. 422). This mirrors the Braille phenomenon noted earlier, where prosthetic distortions may increase separability for certain symbol pairs. 
This suggests that legibility may not degrade linearly with signal fidelity, motivating computational search over intuitive assumptions.

Second, \emph{diversity alone improves legibility.} Despite the occasional resilience of native symbols, randomized assignments from the mixed pool consistently outperform the native baseline (e.g., Arabic Medium: 327 vs.\ 987). This suggests that standard alphabets, constrained by typographic cohesion, are generally suboptimal for prosthetic vision compared to a mixed pool containing distinct shapes like \ac{DCT} gratings and Braille.

Third, \emph{optimization provides the critical gain.} \textsc{SymbolSight} outperforms even the \emph{best} randomized seeds by factors of 3--8$\times$. This confirms that minimizing Eq.~\ref{eq:confusion_cost}, specifically avoiding collisions for high-frequency bigrams, yields gains far beyond what chance selection can provide.

Finally, \emph{the improvement is robust.} Table~\ref{tab:improvement_factor} shows \textsc{SymbolSight} reduces predicted confusion by a median factor of 21.6$\times$ compared to native text (range 12.5--29.6$\times$). 
These values should therefore be interpreted as relative design scores for codebook selection, not as predicted letter-recognition accuracies for current prosthesis users.

\subsection{Qualitative Structure of Optimized Alphabets}
\begin{figure}[!tb]
    \centering
    \includegraphics[width=0.999\linewidth]{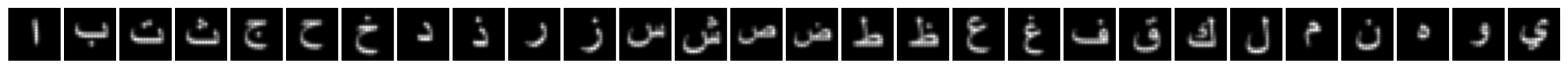}
    \includegraphics[width=0.999\linewidth]{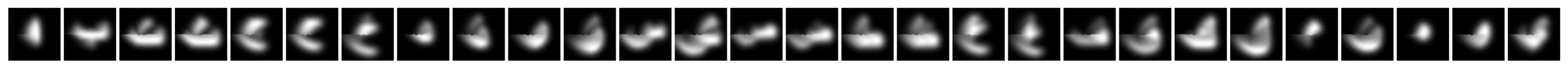}
    \includegraphics[width=0.999\linewidth]{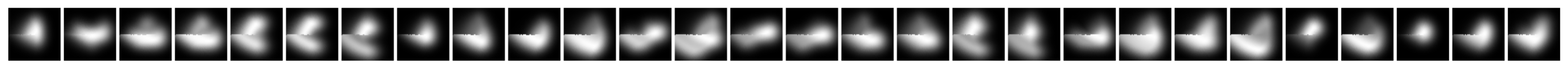}

    \vspace{0.2cm}
    \includegraphics[width=0.999\linewidth]{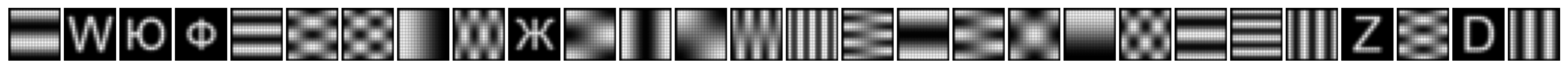}
    \includegraphics[width=0.999\linewidth]{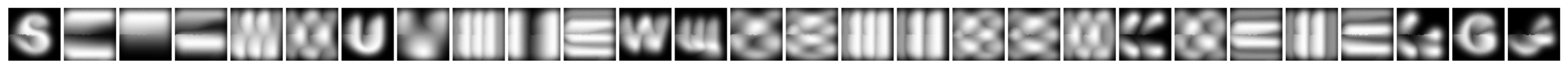}
    \includegraphics[width=0.999\linewidth]{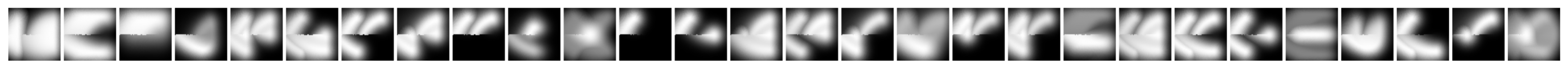}
    \caption{
         \textbf{Arabic symbol comparison.} Top three rows: native Arabic at low, medium, and high distortion. Bottom three rows: optimized symbols at each distortion level in matching order. 
         At high distortion, native characters blur together (Row 3), whereas optimized glyphs retain more distinct structural footprints (Row 6).
    }
    \label{fig:arabic_symbol_rows}
\end{figure}

\begin{figure}[!tb]
    \centering
    \includegraphics[width=0.999\linewidth]{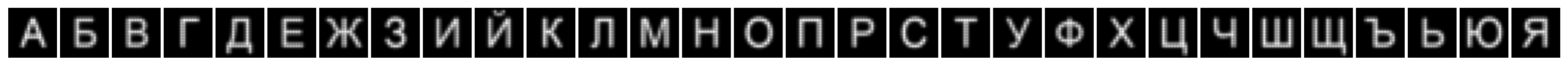}
    \includegraphics[width=0.999\linewidth]{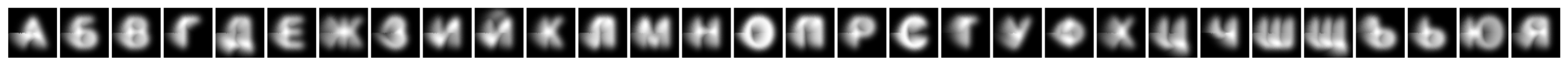}
    \includegraphics[width=0.999\linewidth]{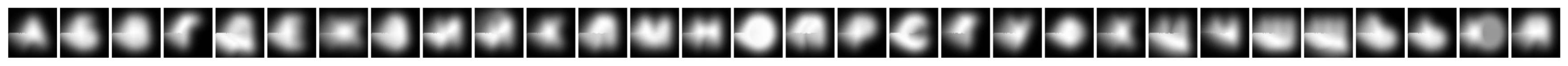}

    \vspace{0.2cm}
    \includegraphics[width=0.999\linewidth]{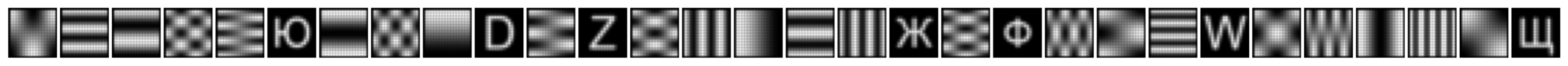}
    \includegraphics[width=0.999\linewidth]{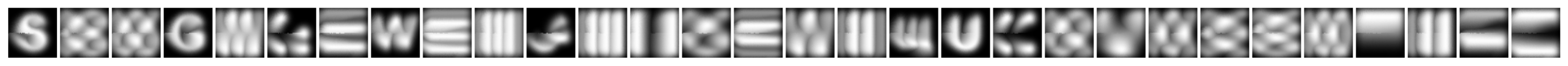}
    \includegraphics[width=0.999\linewidth]{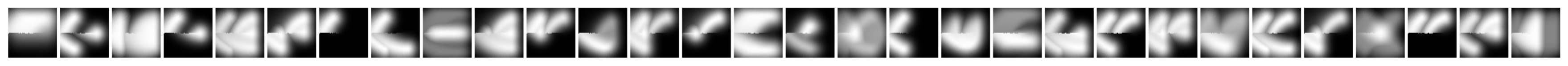}
    \caption{
         \textbf{Bulgarian symbol comparison.} Top three rows: native Cyrillic at low, medium, and high distortion. Bottom three rows: optimized symbols at each distortion level in matching order. 
    }
    \label{fig:bulgarian_symbol_rows}
\end{figure}

\begin{figure}[!tb]
    \centering
    \includegraphics[width=0.999\linewidth]{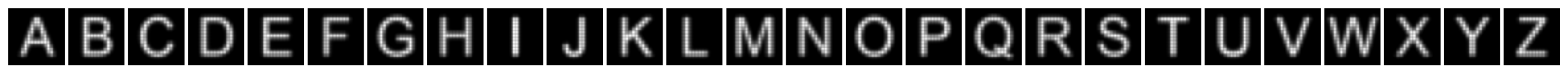}
    \includegraphics[width=0.999\linewidth]{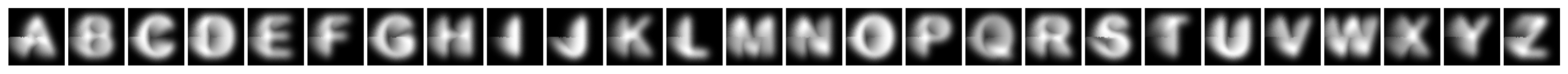}
    \includegraphics[width=0.999\linewidth]{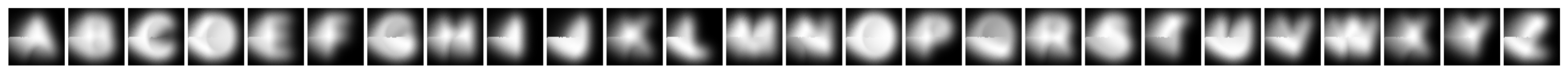}

    \vspace{0.2cm}
    \includegraphics[width=0.999\linewidth]{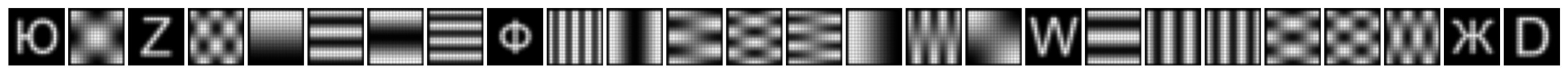}
    \includegraphics[width=0.999\linewidth]{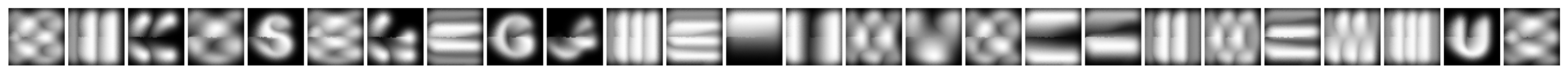}
    \includegraphics[width=0.999\linewidth]{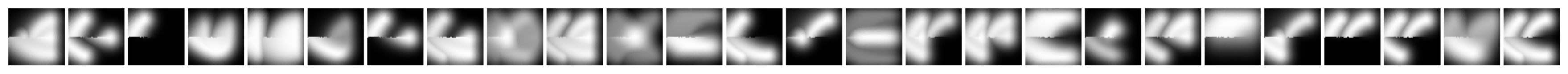}
    \caption{
         \textbf{English symbol comparison.} Top three rows: native Latin at low, medium, and high distortion. Bottom three rows: optimized symbols at each distortion level in matching order.
    }
    \label{fig:english_symbol_rows}
\end{figure}

Figures~\ref{fig:arabic_symbol_rows}, \ref{fig:bulgarian_symbol_rows}, and~\ref{fig:english_symbol_rows} show the symbols selected for each language at each distortion level. Several observations stand out. 
First, the optimized sets are heterogeneous: they draw from Latin, Cyrillic, Braille, and \ac{DCT} sources, mixing alphabets and synthetic patterns within a single assignment. This mixing exploits cross-set distinctiveness visible in the confusion matrices (Fig.~\ref{fig:training_confusion_symbol_sets}), where symbols from different families are rarely confused.

Second, as distortion increases, the algorithm shifts toward symbols with coarse, high-contrast structure, preferring Braille and \ac{DCT} symbols, as their low-frequency content survives blur better than fine strokes.

Third, comparing native symbols at high distortion (third row of each figure) to optimized symbols at high distortion (sixth row) makes the benefit apparent: many native letters become visually inseparable under the simulated distortion, whereas optimized symbols retain clearly different gross shapes.

Finally, the benefit at low and medium distortion is less visually obvious in static images because temporal distortion, the ghosting of sequential symbols, cannot be depicted in a single frame. The confusion cost reductions (Tables~\ref{tab:confusion_cost} and~\ref{tab:improvement_factor}) confirm that optimized sets also outperform baselines in these conditions.

\section{Discussion}

\textsc{SymbolSight} is motivated by the observation that in serial, symbol-at-a-time prosthetic reading, the dominant bottleneck is not single-symbol recognition, but interference from the persistence of the \textit{previous} symbol. By explicitly penalizing visual similarity between frequently adjacent letters (high-probability bigrams), the framework targets the error modes introduced by temporal nonlinearities rather than attempting to reconstruct fine spatial detail. 
The observed 22$\times$ reduction in predicted confusion cost suggests that native alphabets are poorly matched to this modality and that substantial gains may be achievable through software-level optimization.

\subsection{Optimizing the Symbol Codebook Matters}
The central finding of this study is that symbol choice and assignment have a first-order impact on predicted reading performance under prosthetic vision constraints, independent of hardware changes.
Across three languages and multiple distortion regimes, optimized symbol-to-letter mappings consistently outperformed native alphabets and randomized heterogeneous baselines (Table~\ref{tab:improvement_factor}). This indicates that standard typography, designed for high-acuity parallel reading, constitutes a suboptimal codebook for serial, low-bandwidth prosthetic vision, and that heterogeneous symbol sets can substantially reduce temporal interference.

\subsection{Modeling Assumptions and Limitations}
A primary challenge in prosthetic vision research is the combinatorial explosion of parameter spaces, which renders exhaustive human testing impractical. We position this framework not as a replacement for clinical trials, but as a computational sieve to identify high-potential symbol sets.
In current retinal implants, percepts may appear as elongated streaks, connected segments, or partial contours, and users often rely on eye and head movements to accumulate information over time.
Accordingly, the network is used only to rank relative structural separability across candidate symbol pairs.
Our simulation also relies on approximations. Temporal persistence is modeled via MixUp-based linear superposition, which abstracts complex decay dynamics~\cite{hou_predicting_2024} into feature-space interference. 
In addition, the resulting confusion matrices depend on rendering choices such as font weight, size, and stroke geometry. While we standardized these parameters to isolate symbol geometry, alternative typographic instantiations would alter the confusion structure. Importantly, these factors do not limit the framework itself, as both temporal models and typographic parameters can be incorporated directly into the optimization loop.

\subsection{Clinical Interpretation and Path Forward}
While natural reading relies on suprasegmental cues such as word shape, these cues require parallel letter processing and are unavailable in the symbol-at-a-time modality imposed by current low-resolution prostheses~\cite{paraskevoudi_full_2021}. Consequently, the heterogeneity of the optimized symbol sets does not sacrifice word-level information, because temporal sequencing has already eliminated it. From a clinical perspective, the relevant trade-off is therefore not visual cohesion, but reliability.
Although the optimized sets introduce a learned mapping that mixes Latin letters with Braille- and \ac{DCT}-like symbols, prosthetic vision already demands substantial rehabilitation. In this context, learning an optimized symbol code may redirect training toward a more reliable input signal rather than simply adding burden~\cite{rassia_improvement_2018}. 
Clinical usability will depend on whether users can learn and retain these heterogeneous mappings, whether the symbols remain discriminable during active scanning, and whether personalized codebooks outperform familiar alphabets in psychophysical testing. Looking forward, this work can be extended beyond letter-level optimization to word-level encodings and re-instantiated using device- or user-specific phosphene maps~\cite{granley2023human} as these become available for retinal and cortical implants.

\section{Conclusion}

We introduced \textsc{SymbolSight}, which treats sequential prosthetic reading as communication over a noisy channel with memory and optimizes the symbol codebook rather than the hardware. Using \ac{SPV} and a proxy observer to estimate pairwise interference costs, then weighting these costs by language-specific bigram statistics, \textsc{SymbolSight} reduced the expected confusion cost by a median factor of 22$\times$ across Arabic, Bulgarian, and English, consistently outperforming native alphabets and randomized heterogeneous baselines.  In future work, we will validate these findings through human psychophysical experiments under sequential presentation and extend the framework to incorporate device- and user-specific percept maps.

\bibliographystyle{ieeetr} 

\bibliography{main_0068}

\end{document}